\documentclass{article}

\PassOptionsToPackage{numbers, compress}{natbib}
\usepackage{grffile}

    \usepackage[final]{neurips_2023}

\usepackage{pythonhighlight}

\usepackage[utf8]{inputenc} 
\usepackage[T1]{fontenc}    
\usepackage{hyperref}       
\usepackage{url}            
\usepackage{booktabs}       
\usepackage{amsfonts}       
\usepackage{nicefrac}       
\usepackage{microtype}      
\usepackage{amsmath}
\usepackage{natbib}
\usepackage{algorithm}
\usepackage{booktabs} 
\usepackage{caption}
\usepackage{xcolor}
\usepackage[noend]{algpseudocode}
\usepackage{wrapfig}
\usepackage{adjustbox}
\usepackage{subfigure}
\usepackage{changepage}
\usepackage{sidecap}

\def\etal{{\em et al.~}}
\def\ie{\emph{i.e.~}}

\title{Key-Value Transformer}

\author{Ali Borji \\
Quintic AI\\
San Francisco, CA\\
\texttt{aliborji@gmail.com} 
}

\begin{document}

\maketitle

\begin{abstract}

Transformers have emerged as the prevailing standard solution for various AI tasks, including computer vision and natural language processing. The widely adopted Query, Key, and Value formulation (QKV) has played a significant role in this. Nevertheless, no research has examined the essentiality of these three components for transformer performance. Therefore, we conducted an evaluation of the key-value formulation (KV), which generates symmetric attention maps, along with an asymmetric version that incorporates a 2D positional encoding into the attention matrix. Remarkably, this transformer requires fewer parameters and computation than the original one. Through experiments encompassing three task types—synthetics (such as reversing or sorting a list), vision (mnist or cifar classification), and NLP (character generation and translation)—we discovered that the KV transformer occasionally outperforms the QKV transformer. However, it also exhibits instances of underperformance compared to QKV, making it challenging to draw a definitive conclusion. Nonetheless, we consider the reported results to be encouraging and anticipate that they may pave the way for more efficient transformers in the future\footnote{Code is available at \url{https://github.com/aliborji/kv-transformer}}.

\end{abstract}

\section{Introduction}


Transformers (Vaswani~\etal~\cite{vaswani2017attention}) have gained significant attention in recent times due to their effectiveness in various domains such as language, vision, and reinforcement learning. The research community has witnessed a remarkable surge in the number of proposed Transformer model variants. These "X-former" models, including Reformer, Linformer, Performer, and Longformer, to name a few, aim to enhance the original Transformer architecture with advancements in computational and memory efficiency. For a detailed and comprehensive review of this topic, please refer to the review on efficient Transformer models by Tay~\etal~\cite{tay2022efficient}.

To our current understanding, the majority of X-formers have primarily focused on the Query, Key, and Value (QKV) formulation. However, it remains unclear whether these three components are essential or if solely having the key-value aspect is sufficient. Furthermore, the extent to which the performance may be compromised by removing the query component is not well-established. It is still uncertain whether the QKV formulation holds inherent theoretical significance or if it is primarily a matter of parameters. In simpler terms, is the inclusion of queries, keys, and values crucial to the success of these models? In this study, our aim is to investigate this problem.




\section{Key-Value Transformer}


Transformer blocks are known for incorporating several key components, including a multi-head self-attention mechanism, a position-wise feed-forward network, layer normalization modules, and residual connections. Here, we assume the reader has a basic understanding of the overall transformer architecture and specifically direct our focus towards the multi-head attention, which serves as a central component of the transformer's design.


The self-attention mechanism, also known as intra-attention, is a crucial and distinguishing feature of Transformer models. Its purpose is to establish relationships between different positions within a sequence, enabling the computation of a representation for that very sequence. This mechanism has proven to be highly valuable in various tasks such as machine translation, abstractive summarization, and image description generation.


The fundamental concept underlying self attention mechanism is for each element within the sequence to learn the ability to gather information from other tokens present in the same sequence. The operation for a single head can be defined as follows:
\begin{equation}
    A_h = \text{Softmax}(\alpha Q_h K_h^T)V_h,
\end{equation}
where X is a matrix in $\mathbb{R}^{N\times d}$, $\alpha$ is a scaling factor typically set to $1/ \sqrt d$, $Q_h=XW_q$, $K_h=XW_k$, and $V_h=XW_v$ are 
linear transformations applied on the temporal dimension of the input sequence. $W_q$, $W_k$, $W_v$ $\in \mathbb{R}^{d \times d/H}$ are the 
weight matrices (parameters) responsible for the query, key, and value projections, respectively. These matrices project the input X into an output tensor of $d$ dimensions. $H$ represents the number of heads. Softmax is applied row-wise.


The heads $A_1 \cdots A_H$ are computed in parallel, and their outputs are concatenated and passed through a dense layer. The attention matrix $A = QK^T$ is primarily responsible for learning alignment scores between tokens in the sequence. Notably, this formulation involves taking the dot product between each element/token in the query (Q) and the corresponding element/token in the key (K).


\begin{figure}[t]
    \centering
    \includegraphics[width=.8\linewidth]{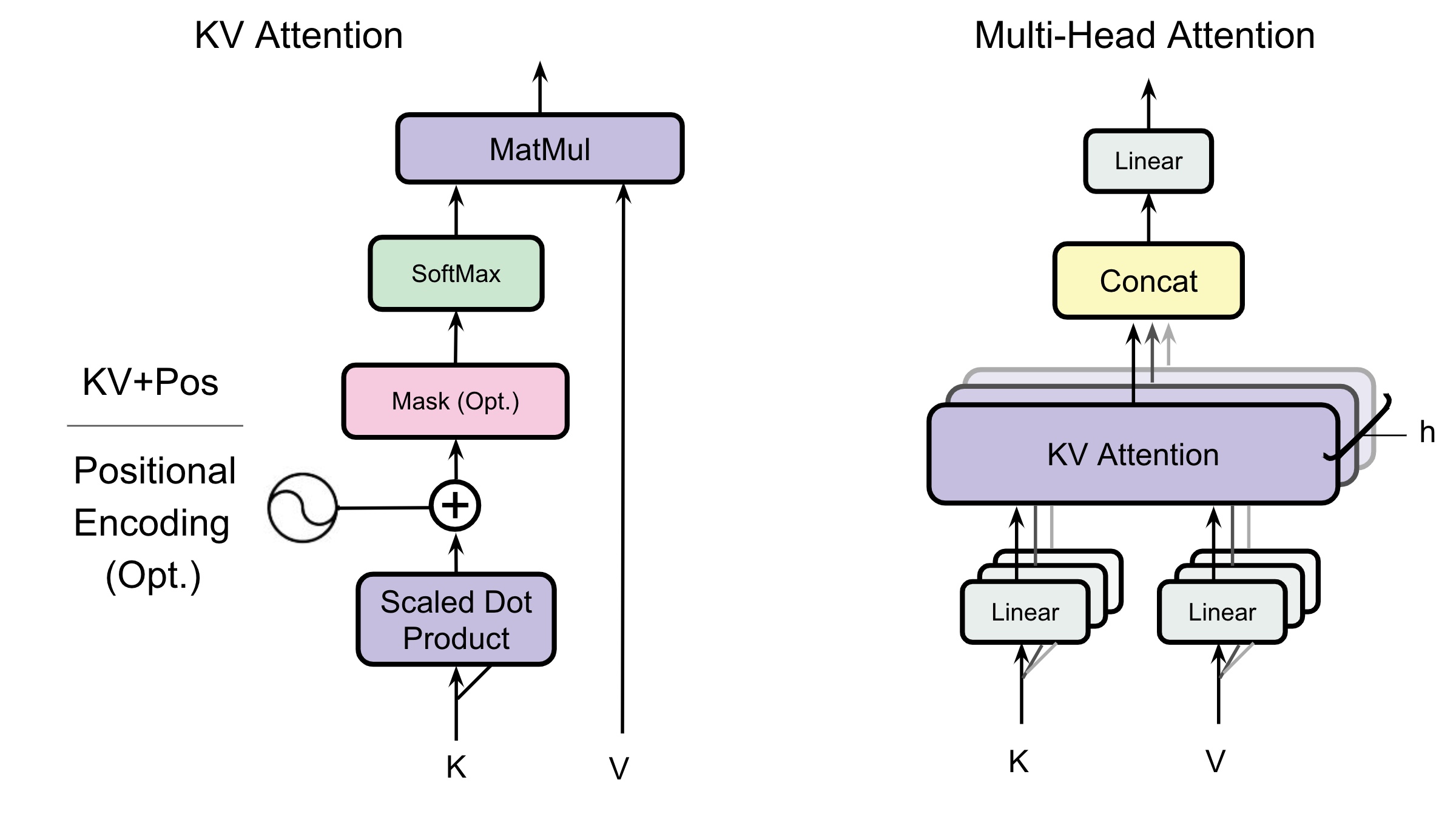}
    \caption{Key-value transformer. The attention with positional encoding is denoted as KV+Pos. In scenarios that require cross-attention, such as translation, we continue to employ the QKV attention mechanism. Removing the Q component is different from weight sharing between Q and K. In our formulation, we completely eliminate Q's parameters and computations.}
    \label{fig:kv}
\end{figure}

In our formulation, we simply replace Q with K, resulting in:
\begin{equation}
    A_h = \text{Softmax}(\alpha K_h K_h^T)V_h,
\end{equation}

As depicted in Figure~\ref{fig:kv}, our approach computes the attention matrix in a manner that maintains symmetry. To introduce asymmetry, a 2D positional encoding of dimension $m$ is added to the $n \times n$ attention matrix\footnote{added to the broadcast attention matrix.}. The resulting tensor, with dimensions $n \times n \times m$, is then projected back into an $n \times n$ matrix using a linear layer consisting of $m$ neurons. In our experiments, we refer to these two variations as KV and KV+Pos attention, respectively.


\begin{table}[htbp]
\centering
\begin{tabular}{c|c|c}
     & Computational Complexity & \# Parameters  \\
  \hline
  QKV &  $2 n d^2$  &  $2 d^2$  \\
  \hline
  KV+Pos &  $n d^2 + n^2m $  &  $d^2 + m $ \\    
  \hline
  KV &  $n d^2 $  &  $d^2$  \\
\hline  
\end{tabular}
\vspace{5pt}
\caption{Comparison of the attention mechanisms in terms of computational complexity and number of parameters. $d$ is the embedding dimensionality, $n$ is the sequence length, and $m$ is the dimensionality of the 2D positional encoding layer (refered to as pos dim or positional dimension across the paper). }
\vspace{-10pt}
\label{tab:kv-spec}
\end{table}


Table~\ref{tab:kv-spec} demonstrates the computational complexity of the two models. For the KV (symmetric) attention, the number of parameters and FLOPS is half of that of the QKV attention. On the other hand, the KV+Pos attention incurs higher parameter and FLOPS counts, which are dependent on the parameter $m$. It is worth noting that $m$ can be set to a much lower value than $d$. However, the drawback of the KV+Pos attention is that its computational complexity is dependent on $n^2$, which can be computationally expensive.

The choice of $m$, whether the addition of positional encoding is beneficial, and whether a symmetric KV attention is sufficient, all depend on the specific problem being addressed. Notice that our formulation provides a trade-off between model complexity and model performance, which becomes particularly important during inference time.

Below is the implementation in PyTorch:

\begin{python}
att = torch.matmul(k, k.transpose(-2, -1))
att /= math.sqrt(self.head_dim)

if add_pos:
    pos = pos_embeddings_2d[:, :T, :T, :] 
    att = att.unsqueeze(-1) + pos.unsqueeze(0)
    att = self.map_pos(att).squeeze(-1)

if apply_mask:
    att = att.masked_fill(self.tril[:T, :T] == 0, float('-inf'))     
att = att.softmax(dim=-1)
\end{python}

It is important to mention that certain tasks, such as translation, may necessitate the use of cross attention. In such cases, we retain the QKV attention mechanism when needed but replace the self-attention with KV attention. Self-attention refers to the scenario where the keys and values are derived from the same source as the queries. On the other hand, in cross-attention, the queries are still generated from the input sequence, but the keys and values are obtained from an external source, such as an encoder module.

\section{Experiments and Results}

We conduct empirical experiments on 13 tasks, exploring various variations of transformers. All models are trained from scratch, except the set anomaly detection. Our objective is not to achieve state-of-the-art performance, but rather to compare the attention mechanisms employed.

Some of the tasks are sequence-to-sequence where the input and the output is a sequence, not necessarily of the same length. Example tasks in this domain is translation. Here a Transformer encoder is used for interpreting the input sequence, and a decoder is used for generating the output in an autoregressive manner. Some other tasks include image classification, list reversal, etc. In these tasks only an encoder is used to map the input to a set of labels.


\subsection{Synthetic tasks}

We focus on five specific tasks outlined below. The input list, which has a predetermined length, consists of numbers ranging from 0 to 9, inclusive of both 0 and 9.

\textbf{Reverse.} In this task, a list of numbers is subjected to a reversal operation. For instance, the input list [4, 3, 9, 8, 1] would be transformed into [1, 8, 9, 3, 4].

\textbf{Sort.} The objective of this task is to arrange the input list in ascending order. For example, [4, 3, 9, 8, 1] would be transformed into [1, 3, 4, 8, 9].

\textbf{Swap.} In this scenario, the first half of the list is exchanged with the second half. For instance, the list [4, 3, 9, 8, 1, 7] would be transformed into [8, 1, 7, 4, 3, 9].

\textbf{Sub.} In this case, each element of the list is subtracted from 9. For example, the array [4, 3, 9, 8, 1] would be transformed into [5, 6, 0, 1, 8].

\textbf{Copy.} In this task, the objective is to retain the input list as is. For example, [4, 3, 9, 8, 1] remains unchanged as [4, 3, 9, 8, 1].










\begin{figure}[t]
    \centering
    \includegraphics[width=.47\linewidth]{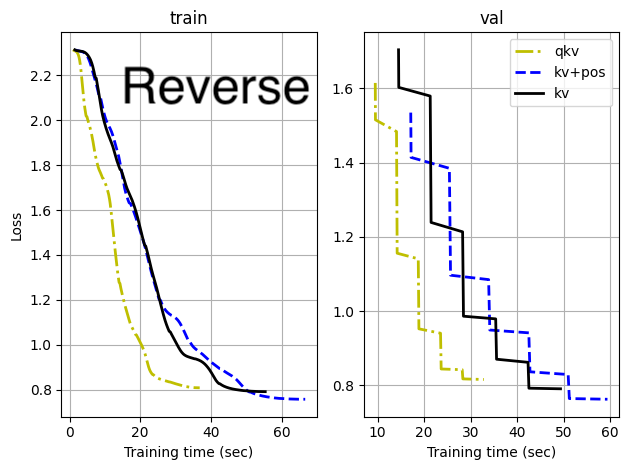}
    \includegraphics[width=.47\linewidth]{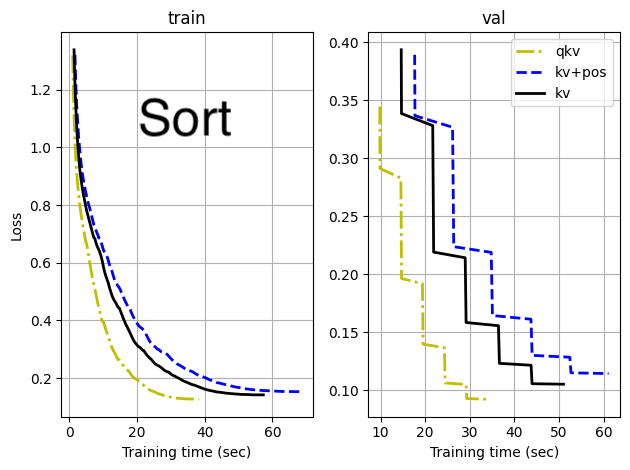}
    \includegraphics[width=.47\linewidth]{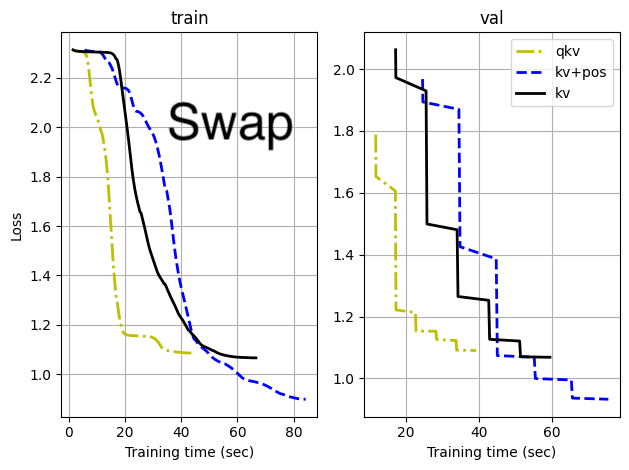}
    \includegraphics[width=.47\linewidth]{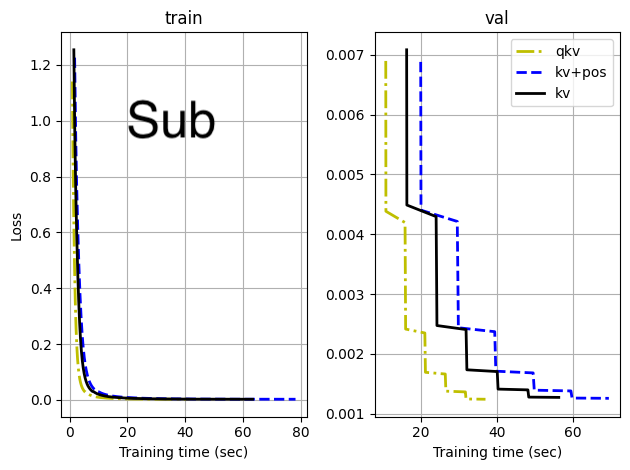}
    \includegraphics[width=.47\linewidth]{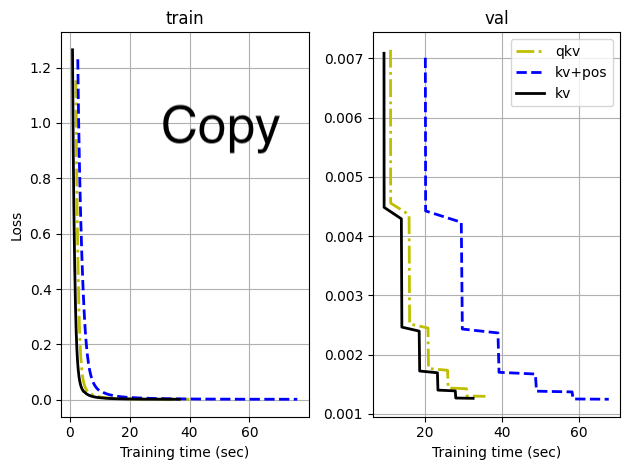}
    \caption{Loss over time for the synthetics tasks.}
    \label{fig:synthetics}
\end{figure}






In training, we feed the input sequence into the Transformer encoder to generate predictions for each token in the input. We utilize the standard cross entropy loss for this purpose. Each number is encoded as a one-hot vector. We apply a gradient clip value of 5 and set the dimensions pos dim to 10 (\ie m). Additionally, we employ the Adam optimizer along with the CosineWarmupScheduler, using a warm-up period of 5.

We conduct experiments with different configurations of the transformer models by varying the embedding dimension (32, 64, 256), the number of layers (2, 4), the number of heads (2, 4), a learning rate of 1e-3, and the input sequence length (16, 64, 128). Each configuration is run three times for two epochs, and the results are then averaged across the variants.

Figure~\ref{fig:synthetics} illustrates the loss curves for both training and validation sets over time. The QKV attention mechanism exhibits faster convergence compared to the KV attention mechanism. However, all attention mechanisms demonstrate good performance on synthetic tasks, as indicated by the accuracies presented in Table~\ref{tab:synthetics}. Notably, the KV+Pos model outperforms other models in this scenario, while the KV transformer model performs slightly better than the QKV transformer model.



\begin{table}[t]
\small
\caption{The performance of attention mechanisms on synthetic tasks. Standard deviations across various architecture hyperparameters (such as model architecture, learning rate, sequence length, etc.) are indicated within parentheses. Multiple runs are conducted for each setting, and the results are averaged across these settings.}
\label{tab:synthetics}
\begin{center}
\begin{tabular}{c|ccccc|c}
 & reverse &  sort & swap & sub & copy & avg.\\
\hline
QKV & 0.698 (0.38) & 0.971 (0.033) & 1.0 (0.0)  & 0.588 (0.42) & 1.0 (0.0) & 0.851 \\
KV+Pos & 0.718 (0.35)  & 0.963 (0.038)  &  1.0 (0.0)  & 0.671 (0.37) &  1.0 (0.0) & {\bf 0.87} \\
KV & 0.705 (0.37)  & 0.967 (0.035)   &1.0 (0.0)  &  0.597 (0.41)  & 1.0 (0.0) & 0.854\\
\hline
\end{tabular}
\end{center}
\end{table}

\begin{figure}[htbp]
    \centering
    \vspace{-10pt}
    \includegraphics[width=.3\linewidth]{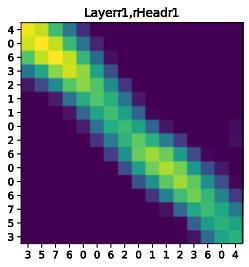}
    \includegraphics[width=.3\linewidth]{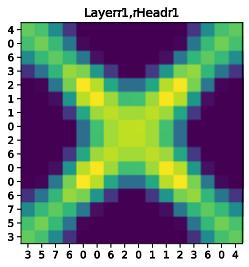}
    \includegraphics[width=.3\linewidth]{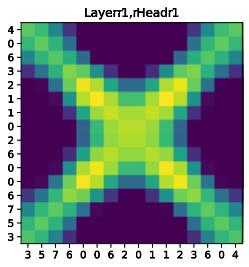}
    \includegraphics[width=.3\linewidth]{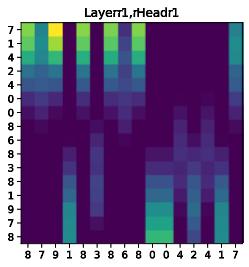}
    \includegraphics[width=.3\linewidth]{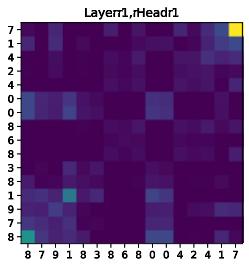}
    \includegraphics[width=.3\linewidth]{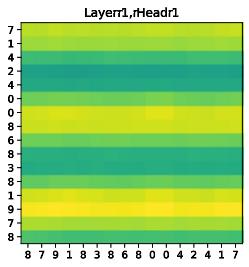}
    \includegraphics[width=.3\linewidth]{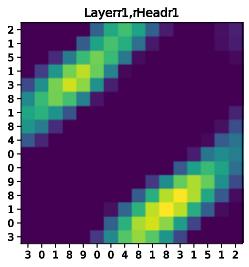}
    \includegraphics[width=.3\linewidth]{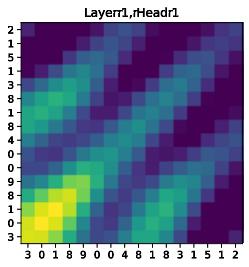}
    \includegraphics[width=.3\linewidth]{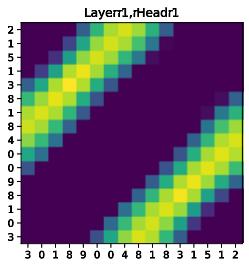}
    \includegraphics[width=.3\linewidth]{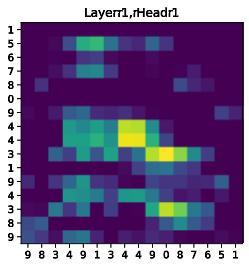}
    \includegraphics[width=.3\linewidth]{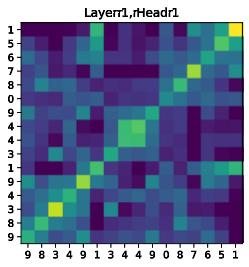}
    \includegraphics[width=.3\linewidth]{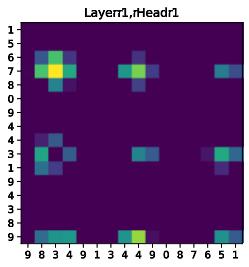}    
    \includegraphics[width=.3\linewidth]{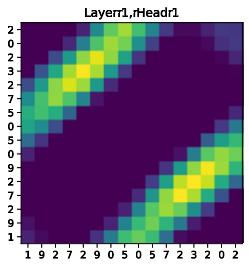}
    \includegraphics[width=.3\linewidth]{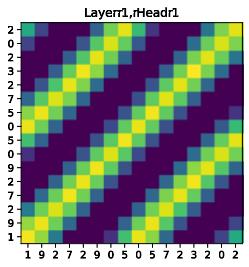}
    \includegraphics[width=.3\linewidth]{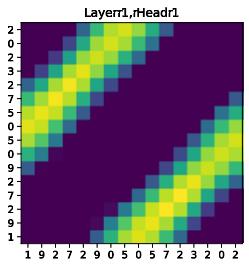}        
    \caption{Attention maps over synthetic tasks. Rows from top to bottom: Reverse, Sort, Swap, Sub, and Copy. Columns from left to right: QKV, KV, and KV+Pos.}
    \label{fig:synthetics_attn}
\end{figure}

Figure~\ref{fig:synthetics_attn} displays sample attention maps. It is worth noting that the attention maps of the KV transformer exhibit symmetry around the line $y=x$. Notable patterns can be observed within the attention maps. For instance, in the reversing task, the QKV model has learned to attend to the token located at the flipped index of itself. However, it also allocates some attention to values near the flipped index. This behavior arises because the model does not require precise, strict attention to solve this problem but rather benefits from an approximate, noisy attention map.


\subsection{Vision tasks}
We consider classification over MNIST, FashionMNIST, CIFAR-10, and CIFAR-100 datasets, as well as finding an anomaly image in a set of images.


{\bf Classification}.
We explore various settings for the patch size (4, 7), learning rate (1e-3, 1e-4), embedding dimension (64, 256, 512), number of layers (2, 4), and number of heads (2, 4). For each configuration, we run two experiments, with each experiment lasting k epochs. The value of k differs depending on the dataset: 20 epochs for MNIST and FashionMNIST, 40 epochs for CIFAR-10, and 50 epochs for CIFAR-100. We employ the cross entropy loss function and utilize the Adam optimizer with the MultiStepLR scheduler for optimization. In the case of 2D positional encoding, we set pos dim to 50.

\begin{figure}[htbp]
    \centering
    \includegraphics[width=.47\linewidth]{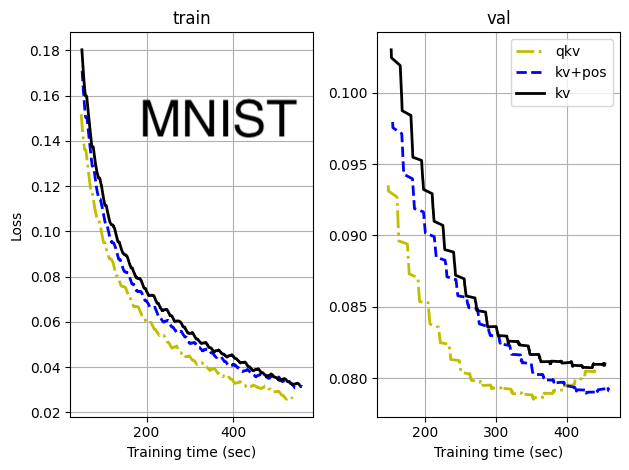}
    \includegraphics[width=.47\linewidth]{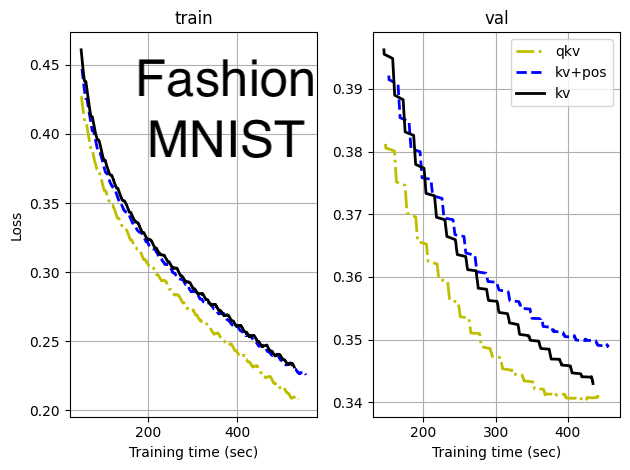}
    \includegraphics[width=.47\linewidth]{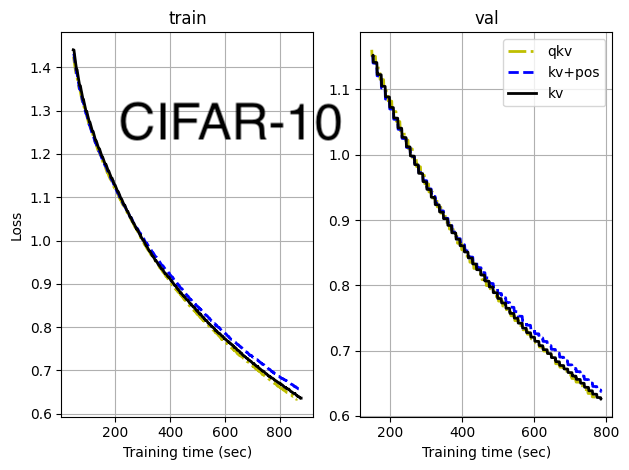}
    \includegraphics[width=.47\linewidth]{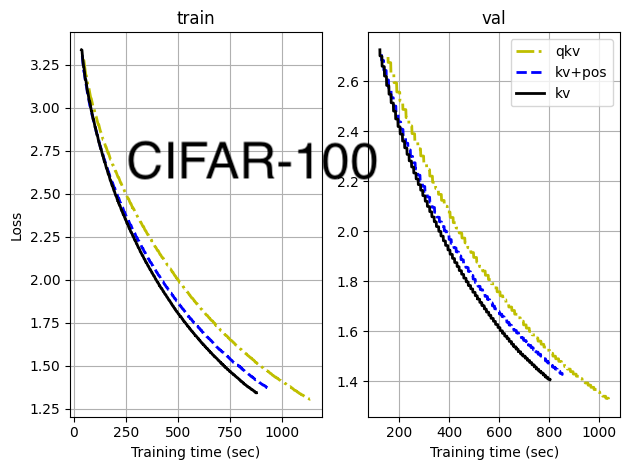}
    \includegraphics[width=.47\linewidth]{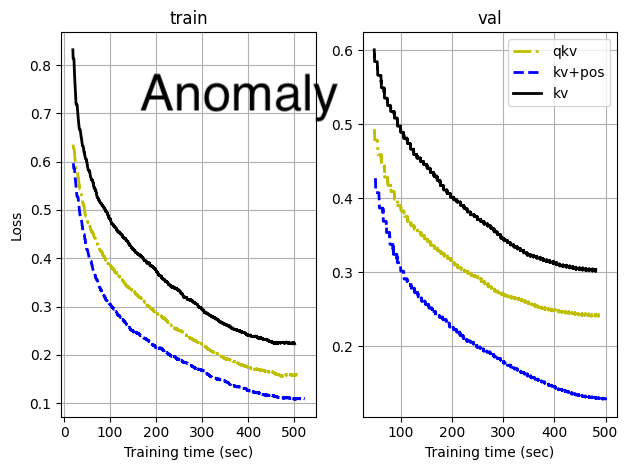}
    \caption{Loss over time for the vision tasks.}
    \label{fig:vision}
\end{figure}

The loss curves are depicted in Figure~\ref{fig:vision}, while the corresponding accuracies are presented in Table~\ref{tab:vision}. It is evident that the KV+Pos attention mechanism achieves superior performance compared to the other variants, with the KV attention mechanism following closely behind, and the QKV attention mechanism performing slightly lower in terms of accuracy.

{\bf Set Anomaly Detection.} 
The CIFAR-100 dataset was utilized, consisting of 600 images distributed across 100 classes, each with a resolution of 32x32 pixels. Our objective is to present the model with a set of 9 images from one class and 1 image from a different class. The task is to identify the image that belongs to the different class among the set. To extract high-level, low-dimensional features from the images, we employ a pre-trained ResNet34 model~\cite{he2016deep} trained on the ImageNet dataset~\cite{deng2009imagenet}. To monitor the training progress and determine when to stop, a validation set is created. In this scenario, we divide the training set into 90\% for training purposes and 10\% for validation, ensuring a balanced distribution across classes.

In our approach, we define an epoch as a sequence in which each image within the dataset is considered as an ``anomaly" exactly once. Therefore, the length of the dataset is determined by the total number of images it contains.

When constructing the training set, we follow a two-step process. First, we randomly sample a class that is different from the class of the image at the corresponding index, denoted as ``idx." Then, in the second step, we sample N-1 images from the newly selected class. This process ensures that the resulting set consists of 10 images, with 9 images belonging to a single category and 1 image serving as the anomaly.





Here, we have a classification of the whole set. For the prediction to be permutation-equivariant, we will output one logit for each image. Over these logits, we apply a softmax and train the anomaly image to have the highest score/probability. This is slightly different than a standard classification layer as the softmax is applied over images, not over output classes in the classical sense. However, if we swap two images in their position, we effectively swap their position in the output softmax. Hence, the prediction is equivariant with respect to the input.




Our transformer model consists of four layers, each equipped with four attention heads. The hidden dimensionality of the model is set to 256, and we incorporate a dropout rate of 0.1 throughout the model to facilitate regularization. 


To control the model's learning rate, we utilize the CosineWarmupScheduler. We configure the warm-up parameter (set to 100) to gradually initiate the model training process.

In our experiments, we vary the embedding dimension, selecting from the options of 256 and 512. Additionally, we explore different depths and numbers of heads, choosing values of 2 and 4. We set the learning rate to 5e-4 for all configurations. Each setting is executed twice for a total of 20 epochs, and the results are subsequently averaged to obtain reliable performance measurements.

\begin{table}[t]
\small
\caption{The performance of attention mechanisms on vision tasks.}
\label{tab:vision}
\begin{center}
\begin{tabular}{c|ccccc|c}
 & mnist &  f-mnist & cifar10 & cifar100 & anomaly & avg.\\
\hline
QKV & 0.981 (0.002) & 0.887 (0.004)  & 0.663 (0.019) & 0.363 (0.016) & 0.943 (0.003) & 0.767\\
KV+Pos & 0.982 (0.002)  & 0.884 (0.006)  & 0.662 (0.017) & 0.366 (0.012) & 0.965 (0.012) & {\bf 0.772} \\
KV & 0.981 (0.002)  & 0.885 (0.005)  & 0.666 (0.021) & 0.369 (0.010)   & 0.927 (0.01) & 0.766  \\
\hline
\end{tabular}
\end{center}
\end{table}



As indicated in Table~\ref{tab:vision}, the KV+Pos transformer exhibits comparable performance to the QKV transformer across the MNIST, FashionMNIST, and CIFAR-10 datasets. However, when it comes to the CIFAR-100 dataset and anomaly tasks, the KV+Pos transformer surpasses the QKV transformer in terms of accuracy, resulting in an overall higher average performance. The KV transformer, while slightly behind these two variants, still performs at a reasonably competitive level.

\subsection{NLP tasks}



\textbf{Character generation.} The objective in this task is to generate text by predicting the next character\footnote{For more detailed information about the model used for this task, please refer to \url{https://www.youtube.com/watch?v=kCc8FmEb1nY&t=6064s&ab_channel=AndrejKarpathy}.}. We utilize the tiny Shakespeare dataset\footnote{\url{https://raw.githubusercontent.com/karpathy/char-rnn/master/data/tinyshakespeare/input.txt}}, which consists of 1,115,394 characters.

\begin{figure}[t]
    \centering
    \includegraphics[width=.6\linewidth]{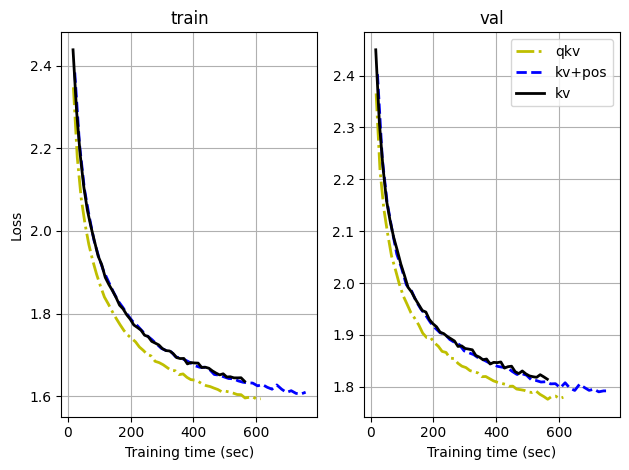}
    \caption{Loss over time for the character generation task.}
    \label{fig:gpt}
\end{figure}

For the training process, we allocate 90\% of the data for training purposes, while the remaining portion is used for validation. During each training step, we track the loss across the entire validation set. Several hyperparameters are considered, including the context size (8, 32, 64), maximum iteration set to 1000, learning rate of 5e-4, embedding dimension chosen from 64 and 192, number of heads selected from 1 and 4, number of layers chosen from 1, 2, 3, and 6, dropout rate set to 0.2, and positional dimension set to 20. The optimization process involves using the Adam optimizer along with cross entropy loss. In total, four models are trained. The results are illustrated in Fig.~\ref{fig:gpt}. It is evident that the KV+Pos and KV transformers perform similarly, while the QKV transformer exhibits faster convergence.

\textbf{Number generation.} In this task, our objective is to generate the next token in a dataset that comprises written numbers. The dataset includes consecutive numbers from 1 to 9999, spelled out as words. The task involves predicting the subsequent token given the preceding $l$ tokens. For instance, if the sequence length is $l=3$, we would have examples like:

(['thirteen', '.', 'fourteen'], '.') \newline
(['.', 'fifteen', '.'], 'sixteen') \newline
$\cdots$

The vocabulary size for this dataset is 30, and there are a total of 63,095 tokens.

To train transformers, we vary the sequence length $l$ from 16, 32, 64, to 128. Additionally, we adjust the learning rate in the range of 0.001 to 0.0001. Other hyperparameters include a positional dimension of 10, embedding dimension of 64, four layers, and eight heads. Each variation is trained five times for 15 epochs, utilizing the cross entropy loss function.

\begin{figure}[t]
    \centering
    \includegraphics[width=\linewidth]{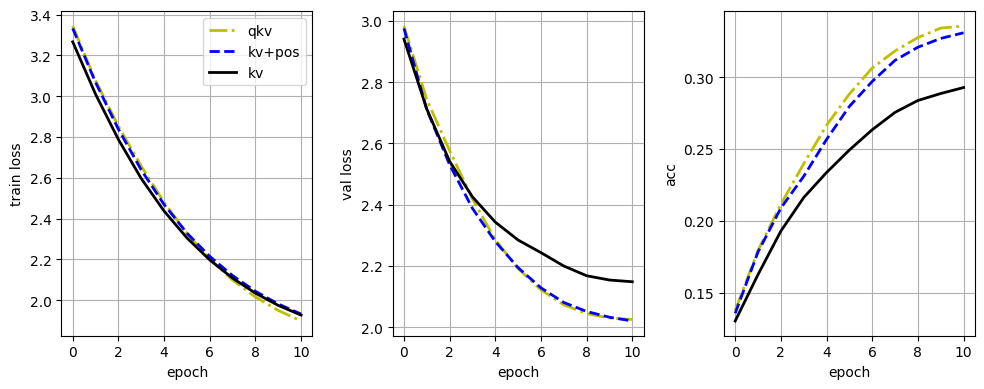}
    \caption{Epoch-wise loss in the number generation task.}
    \label{fig:numbers}
\end{figure}

The results are depicted in Figure~\ref{fig:numbers}. It can be observed that the KV+Pos variant performs closely to the QKV variant, while the KV variant exhibits significantly lower performance.








\textbf{Translation.} 
The objective of this task is to train a transformer model from scratch to perform sentence translation between two languages. Specifically, we utilize the Multi30k dataset to train a German to English translation model as well as an English to German translation model.

In contrast to using a one-hot target distribution, we adopt a different approach. We set the probability of the target word to a predetermined ``confidence value" (typically 0.9) and allocate the remaining ``smoothing value" mass (usually 0.1) evenly across the rest of the vocabulary. This technique, known as label smoothing, aims to provide a more robust training signal.

To optimize the model, we employ the KL divergence loss and the Adam optimizer. The learning rate follows a linear ramp-up for a specified number of warm-up steps (usually 4000) and then decays according to the inverse square root law based on the current training step number. During the translation process, we utilize a greedy decoding strategy, starting with a designated start token.

For the model configuration, we set the positional dimension to 10 and apply a dropout rate of 0.1. We explore different variations by varying the number of layers (1 or 2), the number of heads (1 or 4), and the embedding dimension (64, 128, or 256). Each variant is trained twice for 20 epochs.

\begin{figure}[t]
    \centering
    \includegraphics[width=\linewidth]{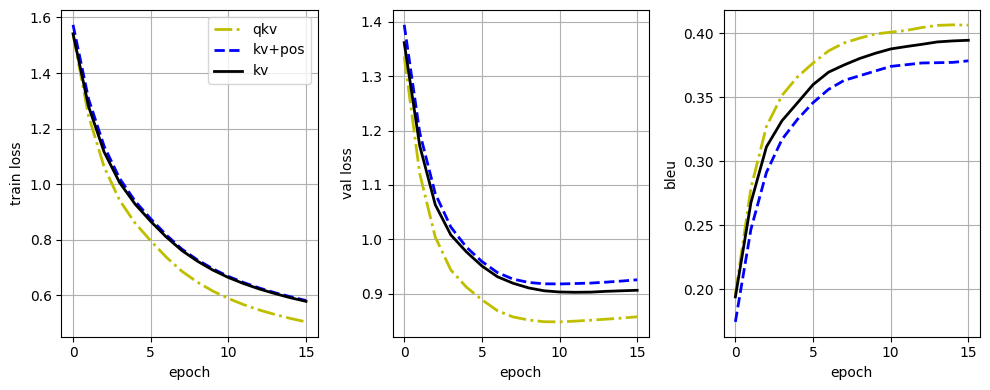}
    \includegraphics[width=\linewidth]{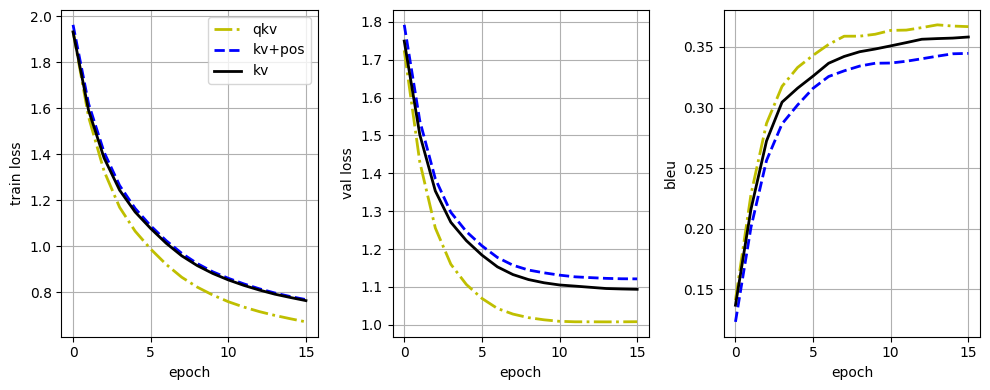}
    \caption{Loss and Bleu per epoch in the translation task. Bleu is computed over the test set. Top: German to English, Bottom: English to German.}
    \label{fig:translation}
\end{figure}

The results for both German to English and English to German translation are depicted in Figure~\ref{fig:translation}. Interestingly, incorporating 2D positional encoding negatively affects the model's performance. However, the KV transformer demonstrates competitive performance compared to the QKV transformer in this task.




\section{Conclusion}

While transformers have brought significant advancements to various domains in AI, our comprehension of self-attention remains somewhat limited. In light of this, we have posed a fundamental question regarding the indispensability of the QKV formulation and endeavored to address it through empirical means. To achieve this, we have put forth two alternative models known as KV and KV+Pos. It is important to note that removing the Q component is distinct from weight sharing between Q and K. In our proposed formulation, we have completely eliminated the parameters and computations associated with the Q component.

We conducted an evaluation comparing the performance of KV attention, both with and without 2D positional encoding, to the commonly used QKV attention across 13 different tasks. Our findings indicate that in certain cases, such as the synthetic and vision datasets, KV attention outperforms QKV attention. It is worth noting that there is a trade-off involved: KV attention (without positional encoding) achieves lower accuracy but requires fewer parameters compared to QKV attention.

Overall, our results highlight the potential benefits of exploring symmetric attention mechanisms. However, it remains unclear how and when symmetry can enhance or hinder model accuracy. To foster further research in this area, we have made the code we used to train and evaluate our models available at: \url{https://github.com/aliborji/kv-transformer}.


\bibliographystyle{plain}
\bibliography{refs}

\begin{thebibliography}{1}

\bibitem{deng2009imagenet}
Jia Deng, Wei Dong, Richard Socher, Li-Jia Li, Kai Li, and Li~Fei-Fei.
\newblock Imagenet: A large-scale hierarchical image database.
\newblock In {\em 2009 IEEE conference on computer vision and pattern
  recognition}, pages 248--255. Ieee, 2009.

\bibitem{he2016deep}
Kaiming He, Xiangyu Zhang, Shaoqing Ren, and Jian Sun.
\newblock Deep residual learning for image recognition.
\newblock In {\em Proceedings of the IEEE conference on computer vision and
  pattern recognition}, pages 770--778, 2016.

\bibitem{tay2022efficient}
Yi~Tay, Mostafa Dehghani, Dara Bahri, and Donald Metzler.
\newblock Efficient transformers: A survey.
\newblock {\em ACM Computing Surveys}, 55(6):1--28, 2022.

\bibitem{vaswani2017attention}
Ashish Vaswani, Noam Shazeer, Niki Parmar, Jakob Uszkoreit, Llion Jones,
  Aidan~N Gomez, {\L}ukasz Kaiser, and Illia Polosukhin.
\newblock Attention is all you need.
\newblock {\em Advances in neural information processing systems}, 30, 2017.

\end{thebibliography}

\end{document}